# Sensing accident-prone features in urban scenes for proactive driving and accident prevention

Sumit Mishra, Praveen Kumar Rajendran, Luiz Felipe Vecchietti, and Dongsoo Har, *Senior Member, IEEE*

**Abstract—** In urban cities, visual information on and along roadways is likely to distract drivers and lead to missing traffic signs and other accident-prone (AP) features. To avoid accidents due to missing these visual cues, this paper proposes a visual notification of AP-features to drivers based on real-time images obtained via dashcam. For this purpose, Google Street View images around accident hotspots (areas of dense accident occurrence) identified by a real-accident dataset are used to train a novel attention module to classify a given urban scene into an accident hotspot or a non-hotspot (area of sparse accident occurrence). The proposed module leverages channel, point, and spatial-wise attention learning on top of different CNN backbones. This leads to better classification results and more certain AP-features with better contextual knowledge when compared with CNN backbones alone. Our proposed module achieves up to 92% classification accuracy. The capability of detecting AP-features by the proposed model were analyzed by a comparative study of three different class activation map (CAM) methods, which were used to inspect specific AP-features causing the classification decision. Outputs of CAM methods were processed by an image processing pipeline to extract only the AP-features that are explainable to drivers and notified using a visual notification system. Range of experiments was performed to prove the efficacy and AP-features of the system. Ablation of the AP-features taking 9.61%, on average, of the total area in each image increased the chance of a given area to be classified as a non-hotspot by up to 21.8%.

## I. INTRODUCTION

ROAD accidents are the prominent cause of deaths and injuries in urban cities. In 2018, the annual projection of deaths worldwide reached 1.35 million [1]. Among the causes of road accidents, road-built conditions as well as high vehicle density are the most prevalent. Traffic congestion makes road traffic chaotic and increases the level of driver stress [2]. Internal and external views of a vehicle perceived by the driver can be cognitively challenging. According to the findings in [2], approximately 14% of driving events occurred where drivers tend to miss critical traffic signs and crucial points of interest (POI) in the road view. With structural complexity of urban cities associated with road structure and the presence of traffic lights, traffic signs, billboards, etc., the volume of visual information is overwhelming [3]. Advisory systems using different modalities of acoustic, visual, and haptic forms have lately been deployed to assuage this problem [4, 5]. These decision support systems have proven to be effective in improving driver behavior and preventing accidents [2].

Many vehicles utilize multiple functions interfaced via a touchscreen on a visual display for assisting drivers. However, the small size of the visual display included in vehicles not only requires a long gaze but also affects driver concentration due to its position with respect to the driver [3]. Depending on location, displays can be categorized into head-down displays (HDDs) and head-up displays (HUDs). Individual effects of HDD and HUD on risky driving and driver concentration have been investigated in previous studies [6]. Despite the popularity of HDDs, they force drivers to take their eyes off the road, hence provoking accident-prone behavior [7, 8]. Conversely, HUDs are harder to implement but more effective, owing to short glance behavior [6], and lead to improved reaction time of drivers. Recently, HUDs have been deployed as smart sun visors [9]. The HUD is typically studied for warning drivers with an apt notification symbol to prevent collision with an anomalous vehicle or pedestrian [3]. The demand for notification of traffic signs and accident-prone (AP) features in the driver's direct line of sight has been studied in [10]. To improve road safety, visual notification systems that improve attention in the driver's line of sight, e.g., using an HUD located in the windscreen, are needed.

Lately, detection of traffic signs with high accuracy by using cameras has been realized with the help of AI techniques. However, detection of AP features, e.g., specific parts of the road scene that might cause an accident, are still a challenging problem. AP features should be detected and actively notified especially near accident hotspots, which are defined as areas with a higher probability of accident occurrence based on historical accident data analysis [11]. In previous studies, accident prediction in an area has been performed considering areal urban element information, weather, areal event information, etc. [12]. The results from these studies can be used by authorities for taking actions to proactively prevent accidents. However, a proactive approach that can warn the

Sumit Mishra is with The Robotics Program, Korea Advanced Institute of Science and Technology, Daejeon, South Korea, sumitmishra209@gmail.com

Praveen Kumar Rajendran is with Division of Future Vehicle, Korea Advanced Institute of Science and Technology, Daejeon, South Korea, praveenkumar@kaist.ac.kr

Luiz Felipe Vecchietti is with Data Science Group, Institute for Basic Science, Daejeon, South Korea, lfelipesv@ibs.re.kr

Dongsoo Har is with The CCS Graduate School of Green Transportation, Korea Advanced Institute of Science and Technology, Daejeon, South Korea, dshar@kaist.ac.kr

supplemental materials for visualization of results and metrics are available at https://github.com/sumitmishra209/APF

Color versions of one or more of the figures in this article are available online at http://ieeexplore.ieee.org

driver of AP road features in real-time is still missing. The implementation of a multi-modal system for detecting AP features in moving vehicles is challenging since the data must be collected in real-time. Proactive approaches for accident prevention have been studied to predict accident occurrence while driving by predicting nearby anomalous vehicle trajectories using dashcam videos [13]. Anomalous trajectory is predicted in [13] by curtailing static features to avoid complexity and interference with dynamic features; hence, structural AP features in the road-scene are not taken into consideration in such proactive systems. In [14], satellite and airborne remote sensing images are used to analyze road safety. However, the satellite images must be downloaded by the vehicle from the data server, which is constrained by real-time availability and internet dependency. Moreover, if road-built AP features are identified in the overhead/top view of the urban physical environment, the driver might find it difficult to identify the same features in planar view observed while driving. Therefore, in addition to accident prediction, it is important to identify and highlight features that can cause accidents in the driver's view.

With the advent of web map services, street-level imagery tools of urban physical environments are readily accessible. These street images contain the fine gradient urban details captured from the earth's surface perspective similar to the viewpoint of drivers. These details include views of neighboring facilities and amenities, house facades, signs, and roads. Street view images also capture the road layout of the streetscape [15]. Therefore, this modality is effective to understand human dynamics in socioeconomic environments and to extract AP features that can be informative to drivers. Leveraging the advantages of street view imagery, studies have been conducted to classify urban elements [16]. Features of urban structure are learned by convolutional neural networks (CNNs) for classification of street-level images [17], street-level mobility prediction [15], street level accessibility [18], building classification [19], road-scene vehicle detection [46], and investigating land-use [20]. For architecture modification to beautify urban spaces, generative adversarial networks are used with street view imagery [21].

Low level features in street view images associated with a higher probability of accident occurrence can be of different size, position, and texture, with no distinct boundaries. These features emerge from textures, colors, and gradients in images and can be a part of AP features. AP features are rather intuitive than discreet for human understanding e.g., the road space where the field of view is occluded by big vehicles. Also, complex parts of an image consisting of main road areas along with adjoining streets near a park or road areas prone to jaywalking from where pedestrians or vehicles may merge to the main road can be AP features. In [22], an unsupervised learning technique is used, using satellite images to extract design features at intersections and which are then classified to find specific design features of an accident-safe road. The complexity of linking these low-level features, within a given image, to accident proneness is rather intuitive. Therefore, to develop a robust AP feature detection system, AP features must be learned from actual hotspots in an automatize fashion for the city of interest.

To achieve this, we propose a model that is trained to classify a given urban scene into an accident hotspot or a non-hotspot (area of sparse accident occurrence). We then identify the features (portions of images) using class activation maps (CAMs) that lead to a scene-view to be classified as an accident hotspot. Those portions can be taken as AP features. The proposed novel attention mechanism that makes the attention maps and attention vectors in the training process weigh the spatial, channel and point features. This attention-based activation can characterize the target area more appropriately. Traditional CNN are first used to extract low-level features. Different attention maps based on these features are then used to activate only important features. This attention-based selection of features reinforces traditional CNN features to improve distinguishing high-level contextual features as accident-prone for an accident hotspot class of scene-view images.

Overall, in this paper, an attentive driving scheme that identifies AP features in urban scenes is proposed. Our attentive driving scheme makes use of CAMs for the feature extraction in conjunction with the CNN based attention model. Our model is trained with a dataset of street view images created from data provided by web map services and can be used in real-time for classifying street view images obtained from a dashcam. CAM is used to understand the classification result by highlighting the relevant portion of an image. A visual notification layout is presented for notifying the detected AP features to the driver using an HUD.

The main contributions of this work are as follows.
- This work firstly introduces the use of ground level AP features extracted from street view images for proactive prevention of driving accidents by visual notification in an HUD. To identify AP features, a methodology that combines a large database of real-accident records and street view images is established.
- Explicit and accountable accident-prone regions of given street view images are obtained by CAM methods using features detected by a novel attention module that is a part of the classifier trained to detect accident hotspots. Point, channel, and spatial attention maps are generated to activate the relevant low-level features for improved high-level features.
- A detailed analysis shows that the proposed method can classify a given area into an accident hotspot with an accuracy up to 92%, as compared to 78% obtained using satellite images only [14]. An ablation study to evaluate the relevance of detected AP features was conducted. The ablation of AP features shows that removing only 9.6%, on average, of the total area in an image containing AP features increases the chance of a given area to be classified as a non-hotspot by up to 21.8%. Therefore, AP features detected by the proposed scheme can be considered as important clues to avoid accidents.

- A novel visual layout design methodology for an attentive driving scheme based on visual notification in an HUD is presented through simulations. To show the viability of the proposed notification system, saliency maps of visual-gaze attention are drawn using [39]. A comparative analysis of the ratio for intersection of saliency map shows that the AP features selected by the proposed model are the ones with the least visual-gaze attention by the driver.

The remainder of this paper is as follows. Section II reports relevant literature. Section III outlines the methodology for data collection and training CNN models to identify AP features. In Section III, a methodology for the visual attentive driving is presented. Section IV presents effectiveness of different CNN classifiers and utilization of CAMs combined with the image processing pipeline along with a MATLAB simulation of visual notification. Section V concludes this work.

## II. LITERATURE REVIEW

The occurrence of accidents is influenced by several factors such as driver's behavior, missing traffic signs, vehicle type, speed, traffic conditions, weather conditions, and road-built structures [12]. Road-scene views and road-built structures are prominent factors of accidents that can be captured when analyzing street view images [11, 23, 24]. Given historic accident data collected and maintained by different city council agencies, it is possible to determine accident hotspots in cities, i.e. locations with dense occurrence of accidents. In previous literature, various clustering algorithms have been proposed and deployed to identify hotspots, including the K-Means algorithm [25], Poisson-Tweedie algorithm [26], Bayesian algorithm, Firefly clustering algorithm [27], Kernel Density Estimator, and DBSCAN algorithm [28]. In this paper, because of its effectiveness proven in similar applications [29], the DBSCAN algorithm is chosen for identifying accident hotspots [2, 28, 30]. Accidents that are members of a cluster are classified as events that occur in a hotspot and the events which do not belong to any cluster are marked as events in a non-hotspot.

Based on the location of hotspots and non-hotspots, street view images are gathered and an attention-based module is combined with a CNN backbone for training a hotspot/non-hotspot binary classifier. Various CNN architectures have been proposed in the literature. In [31] an image classifier, a very deep convolutional network architecture, VGG16 of the visual geometry group is proposed to extract features at low spatial resolution. Squeezenet, a resource-efficient deep neural network with fewer parameters, thus fitting into less memory without sacrificing accuracy, is proposed in [32]. DenseNet proposes a feed-forward connection of each layer such that feature maps of all previous layers are used as inputs of the next layer [33]. Resnet-18 is proposed to skip connections to eschew vanishing gradients and accuracy saturation in deeper architectures [34]. These established CNN architectures have shown high performance in various image processing challenges.

To identify which features, detected from CNN backbones, should be given more importance for a given image, attention mechanisms [65] might be of crucial importance. Attention applied on point, spatial, and channel levels stimulate the detection of important low-level and high-level features. To leverage the use of attention mechanisms to generate more contextual features, we propose a module with channel-wise, spatial-wise [66], and point-wise attention [67]. The proposed model is added to a pre-trained CNN backbone to create the architecture used for the classification task considered in this paper. Refining pre-trained models has been leveraged in previous research for training CNNs using street view image datasets [15, 19, 20]. Similarly, in this paper, due to the limited size of our dataset, we use a pre-trained CNN backbone to train our binary classifier.

Recently, various post-hoc explainable methods have been proposed (CAMs [37], IG [62], LIME [63], SHAP [64]). In our work, we chose to investigate CAMs as they are prevalent in previous studies when combined with CNN architectures. CAMs highlight regions in images that contain features affecting the decision taken by the classifier [37]. Different CAM methods lead to different regions being highlighted as important features for the decision. Gradient-weighted CAM GradCAM++ leverages gradient information of the specific target class to compute the target class weights of each feature map in the last convolution layer of the CNN architecture [36, 37]. Analysis of AP features presented in this paper is performed using three types of previously well-established CAMs: GradCAM [36], GradCAM++ [37], and ScoreCAM [38].

## III. METHODOLOGY

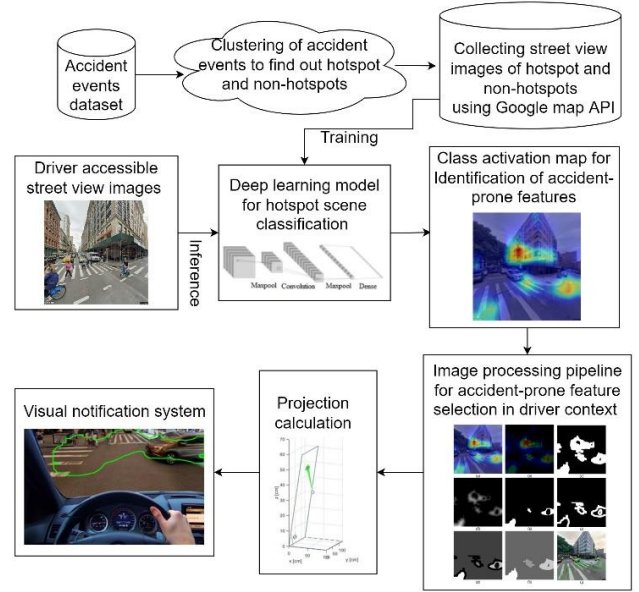

**Fig. 1.** Visual notification for attentive driving.

Figure 1 shows the entire visual notification system including CNN, CAM, the image processing pipeline for feature selections, and visual notification for an HUD on the windshield panel of a vehicle. This section explains the details of the visual notification system.

Here, we first describe the methodology to identify AP features. Hotspots are identified based on historical real accident data and street view images are collected for hotspots and additional images are also collected for non-hotspots to obtain a balanced dataset. Next, we present how the CNN backbone is combined with an attention module to create the proposed architecture and CAM methods for the detection of visual AP features. This section presents the methodology and layout for the visual notification.

### A. Detection of AP features by deep learning models

#### 1) Accident hotspot identification using real accident data of New York city:

Accident data from 2017 to 2021 provided by the department of transportation (DOT) of New York city is used for analysis of motor vehicle collisions, as in [60]. All the accident cases are considered in the analysis. The severity of accidents is not considered, so all the metadata is dropped except for the location information (latitude, longitude) of the accident event. The total number of accident entries in the dataset are 775,443 and cover a vast area with diverse urban structures and views.

The DBSCAN algorithm is used for clustering accident data according to location and requires two hyperparameters: epsilon and minPoints. Clusters are represented as hotspots. The epsilon represents the radius of the circle considered around each data point to check if the density inside the circle and minPoints represents the minimum number of data points required inside that circular region to be identified as a cluster. For finding the optimal epsilon value of the DBSCAN algorithm, the K-Distance graph is used. The optimum epsilon value is the maximum curvature in the K-Distance graph and taken to be 0.0003. The approximate metric equivalent for 0.0001 latitude degree is 11.1 meters, so for 0.0003, the center of the cluster is anywhere within about 60 meters. For minPoints, an empirical value of 50 is taken, meaning that at least 50 accidents should happen for a location to be considered a hotspot. In the New York city data, using these values for the hyperparameters, 2549 clusters (hotspots) are obtained. For accident event entries that are not a part of a cluster, the DBSCAN algorithm marks them as noise using the label '-1' and are considered non-hotspots which accounts for 527,074 accident events.

#### 2) Collecting street view image data for training classifiers and identifying AP features:

To identify AP features, street view images of hotspots and non-hotspots are necessary. Images of non-hotspots are required as samples of a negative class to balance the dataset used for training. To capture street images, the Google Street View API [61] is used. The Google Street View API provides street view imagery as a service with a free quota for the use of API. The images returned from the Google street view API are panoramas with unique IDs. Images can be retrieved by the location coordinate or by the corresponding panorama ID and an angle of view. For an image of a location, the street view image covers an angle of view of approximately 120 degrees.

As per availability, the same panorama is returned by the API if two locations are close. The location of the center of a hotspot is defined here as the average location of the event entries in a cluster. The spread of accident events around the center of a hotspot is limited approximately within the circle of a 60 meter radius. The center location is used for the collection of street view images of the corresponding hotspot. The images are captured in a way to cover roadsides as well as the roadway; therefore, two images are collected for a given location to build the training data: one for +60 degrees and another for -60 degrees set as the angle of view. The size of street images retrieved from the API is 640x640 pixels. A total number of 5,088 images belonging to hotspots are collected. For non-hotspot images, the location of accident events occurred outside any cluster, and thus labeled as noise by the DBSCAN algorithm. A total number of 4,908 image samples corresponding to non-hotspots are used to make a balanced image dataset. The considered hotspots and non-hotspots are presented in Fig. 2.

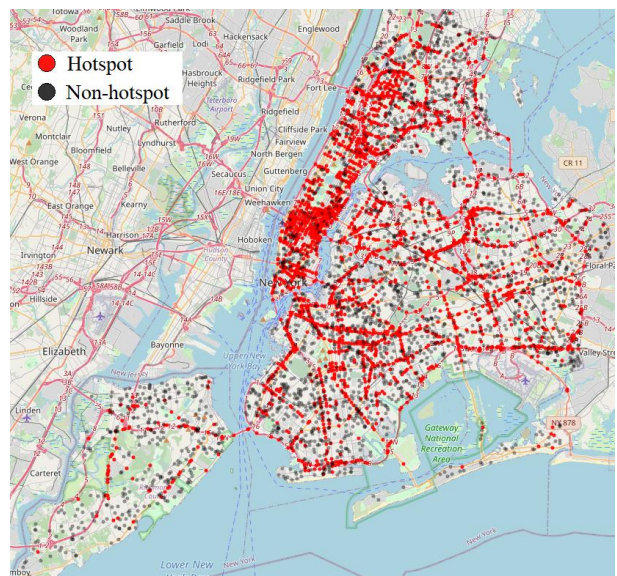

**Fig. 2.** Location of accident hotspots and non-hotspots obtained by using the New York city dataset. Areas without accidents can also be classified as non-hotspots.

#### 3) Image classification using a CNN backbone combined with an attention based module:

Deep learning models, such as CNNs, have achieved breakthrough results while eliminating the need to manually design important features for different tasks such as recharging a sensor network [35, 44], power grid operation [40], and robotic control [41]. CNN models learn visual features that are used to understand an image by utilizing stacked layers of convolution operators and pooling layers to get semantic interpretation. The complexity and interpretation of automatically extracted features evolve layer by layer. The initial layers in a CNN architecture learn low-level features while the final layers in deep CNN architectures learn high-level features for classification [42]. Here, to further characterize these learned features we combine a CNN backbone with an attention-based module (ABM).

The ABM selects regions of the feature vector with the goal of extracting the most significant features while suppressing the insignificant ones. Inspired by [65-67] the ABM (Fig. 3) includes a spatial-wise, channel-wise, or pixel (or point)-wise attention block, which extract relevant contextual information from different levels of abstraction. Channel-wise attention (CA) computes the attention map by exploring the inter-channel relationship among the input features to highlight distinguishable information in different channels. The spatial-wise attention (SA) map is generated by exploring the inter-spatial relation among features. Relevant and similar features across the channel will be highlighted regardless of a long-range spatial gap. Pixel-wise attention (PA) explores each pixel based on its local information. Each module concentrates on identifying relevant features for the task at different levels of abstraction. As shown in Fig. 3, PA is directly given at the output; however, CA features are used to generate SA features. A skip connection from the output of the CNN backbone to the output is also included to preserve low-level features. A detailed study (TABLE I) on the proposed architecture is performed for the binary classification tasks of recognizing hotspots. The best architecture (case (c) in TABLE I) is selected, as presented in Fig. 3.

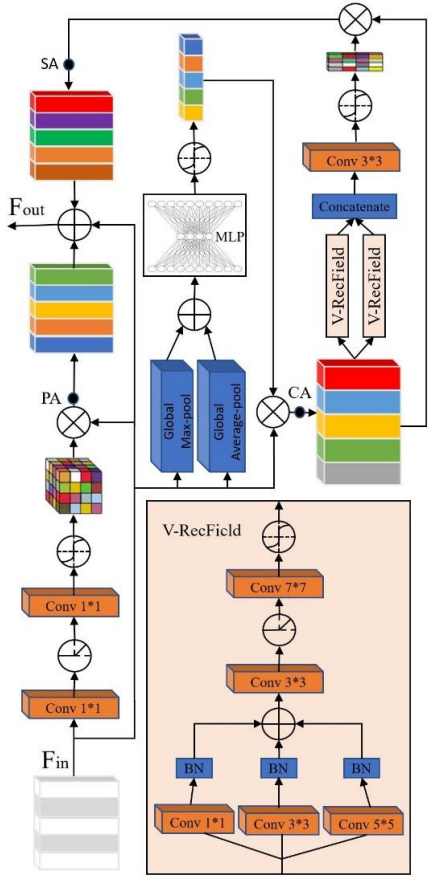

**Fig. 3.** Attention-based module (ABM).

For improved intuition on the ABM architecture, let us assume an input feature map $F_{in} \in R^{H \times W \times C}$ generated from the backbone CNN where H, W, and C represent the height, width, and channels, respectively. For channel attention, spatial information is aggregated by performing global average and global max pooling on the height and width axes to generate $F_{avg}^c \in R^{1 \times 1 \times C}$ and $F_{max}^c \in R^{1 \times 1 \times C}$, respectively. Both $F_{avg}^c$ and $F_{max}^c$ are element-wise added and then passed to a shared network comprised of a multi-layer perceptron (MLP) with one hidden layer of dimension $R^{1 \times 1 \times C/n}$ and ReLu activation function, where n is the compression ratio. Finally, the output of the MLP is passed through a sigmoid activation function $\sigma(.)$ to generate the channel-wise attention map $C_a \in R^{1 \times 1 \times C}$, as shown in (1). The (CA) is then calculated using point wise multiplication of $C_a$ with $F_{in}$, as in (2).

TABLE I

ABLATION STUDY FOR ABM ARCHITECTURE WITH DIFFERENT CNN BACKBONES. DIFFERENT ARCHITECTURES OF ABM ARE PROPOSED FOR STUDY- (a) WITHOUT ABM (ONLY CNN BACKBONE) (b) ABM HAVING ONLY (SA, PA) COMPONENTS AT THE FINAL OUTPUT (c) ABM HAVING (SA, PA, SKIP-CONNECTION) COMPONENTS AT THE FINAL OUTPUT (d) ABM HAVING ALL (SA, PA, CA, SKIP-CONNECTION) COMPONENTS AT THE FINAL OUTPUT.

| CNN Back-bone | ABM Block | Accuracy | Precision | Recall | F1-Score |
|---|---|---|---|---|---|
| Squeeznet | (a) | 0.891 | 0.902 | 0.859 | 0.864 |
| | (b) | 0.890 | 0.903 | 0.855 | 0.863 |
| | (c) | 0.892 | 0.898 | 0.876 | 0.871 |
| | (d) | 0.880 | 0.883 | 0.876 | 0.864 |
| Resnet | (a) | 0.900 | 0.912 | 0.892 | 0.889 |
| | (b) | 0.897 | 0.903 | 0.865 | 0.868 |
| | (c) | 0.900 | 0.915 | 0.869 | 0.875 |
| | (d) | 0.902 | 0.933 | 0.851 | 0.876 |
| VGG | (a) | 0.904 | 0.916 | 0.888 | 0.887 |
| | (b) | 0.904 | 0.908 | 0.866 | 0.872 |
| | (c) | 0.902 | 0.926 | 0.867 | 0.880 |
| | (d) | 0.906 | 0.933 | 0.876 | 0.890 |
| Densenet | (a) | 0.905 | 0.903 | 0.876 | 0.877 |
| | (b) | 0.899 | 0.918 | 0.860 | 0.873 |
| | (c) | **0.922** | **0.947** | **0.901** | **0.913** |
| | (d) | 0.893 | 0.915 | 0.842 | 0.858 |

$$C_a(F_{in}) = \sigma\left(MLP\left(F_{max}^c + F_{avg}^c\right)\right) \quad (1)$$

$$CA = C_a(F_{in}) \otimes F_{in} \quad (2)$$

$$F' = BN(Conv_{1*1}(CA)) + BN(Conv_{3*3}(CA)) + BN(Conv_{5*5}(CA)) \quad (3)$$

$$V = \sigma\left(Conv_{7*7}\left(ReLu(Conv_{3*3}(F'))\right)\right) \quad (4)$$

Next, using CA for spatial attention two parallel varied receptive field (V-RecField) features of a block are concatenated to give $R^{H \times W \times C/2}$. Two parallel blocks help learning variations of the same features. In the V-RecField block for varying receptive field, three different kernel sizes of convolution are used along with a batch-normalization layer, as shown in (3). Furthermore, the feature $(F')$ of the V RecField block is convolved twice by using a ReLu activation in between. Finally, the V-RecField block outputs V after passing through sigmoid activation, as in (4). The concatenated feature vector $(V_1; V_2)$ is convolved again and passed through sigmoid activation to obtain a spatial attention map $(S_a \in R^{H \times W \times 1})$, as in (5). Spatial attention (SA) is then calculated using point wise multiplication of $S_2$ with $F_{in}$, on each channel C, as in (6).

$$S_a(\text{CA}) = \sigma\big(\text{Conv}_{3^*3}(V_1; V_2)\big) \qquad (5)$$
$$SA = S_a(\text{CA}) \otimes F_{\text{in}} \qquad (6)$$

For point attention features, the input $F_{\text{in}}$ is convoluted twice ($\text{Conv}^1_{1\times1}$ and $\text{Conv}^2_{1\times1}$) with an ReLu activation in between. The output of $Conv^2_{1\times1}$ is passed through a sigmoid activation to generate the pixel-wise attention map $P_a \in R^{H \times W \times C}$, as in (7). $P_a$ is then element-wise multiplied with $F_{\text{in}}$ across the spatial and channel axes to generate point attention across the spatial and channel axes to generate the point attention feature (PA), as in (8). The final output of the ABM block is $F_{\text{out}}$, as in (9).

$$P_a = \sigma(Conv^2_{1\times1}) \qquad (7)$$
$$PA = P_a(F_{\text{in}}) \otimes F_{\text{in}} \qquad (8)$$
$$F_{\text{out}} = F_{\text{in}} + PA + CA + SA \qquad (9)$$

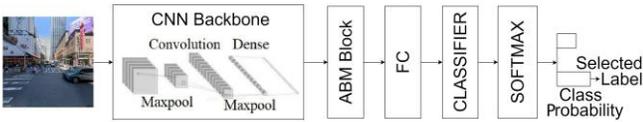

**Fig. 4.** Generic layer-wise architecture of the deep-learning model for classification of hotspot and non-hotspot images.

ABM combined with a CNN backbone is used for classification using the collected image dataset. After obtaining the output from ABM, a fully connected multi-layer perceptron with two outputs is set as the final layer to give a normalized probability score of the two possible classes, hotspot and non-hotspot, using a Softmax layer, as shown in Fig. 4. To identify the input image class, the output of the Softmax layer taking maximum probability is taken as the output class, i.e., maximum probability corresponds to a hotspot or non-hotspot. The CNN backbone is pre-trained for the 1,000-class Imagenet dataset [58] image classification task. These pre-trained models can extract important features for tasks of object identification in images and require less training time with better generalization and a smaller chance of overfitting problems when compared to training the entire network from scratch.

4) *CAMs for sensing visual AP features:*
Visual features associated with an accident are complex and subtle enough to differentiate manually. Identifying static and dynamic features of urban cities and connecting them to an accident have been proven difficult. For example, as a dynamic feature, the road area just behind large vehicles like buses or trucks can act as an accident feature, because the driver's field of view gets occluded. As another example, a wide curve that allows dangerous high-speed driving behavior or a steep curve that is obstructing road view can be static AP features. If these features are identified in real-time by a sensing system and notified to the drivers as hotspots, accidents can be prevented.

Although these features are difficult for humans to identify, deep-learning models are able to identify complex associations of low level and high-level features [42]. Therefore, if the decision on classification by the CNN is known, the features that led to that decision can be analyzed. To analyze the features associated with the CNN classification, CAMs can be used. CAMs are particularly useful to understand the black-box nature of the prediction made by a deep learning model. The CAM represents the weighted sum of high-level features used to identify regions of an image that are being used to identify a given class. Usually, the high-level features considered in the CAMs are those corresponding to the weights of the last convolution layer of the CNN architecture, e.g., the layer before the fully connected layer used for classification. For different CAMs, the weights are considered differently for different features. The CAM-based methods have proven to be viable for understanding and interpreting the model classification in several tasks [45]. In this paper, different CAMs are studied to identify visual AP features that led to the classification of an image as an accident hotspot. These features are processed and displayed in an HUD located in the windshield panel for the driver using the visual notification system.

5) *CAM and related image processing:*

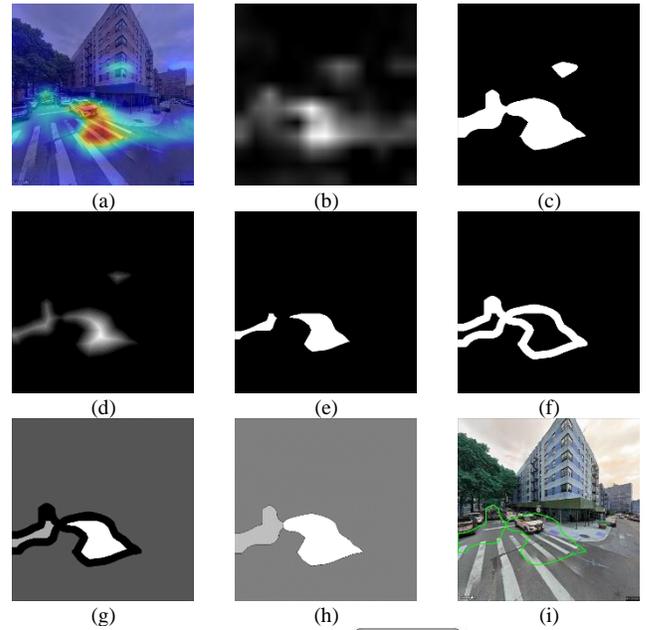

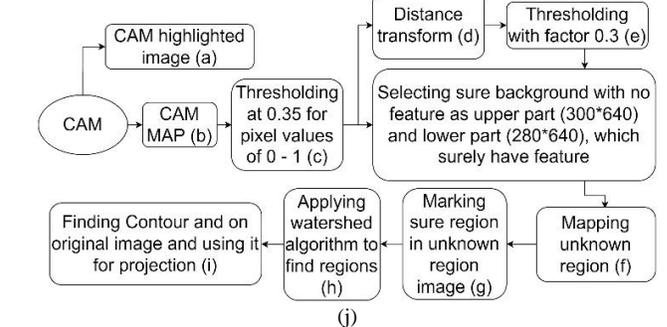

**Fig. 5.** (a-i) Image processing pipeline for selective choice of CAM highlighted features, and (j) image processing pipeline with each block matching with each process labeled with (a-i).

For accident prevention, notification of every feature highlighted by the CAM methods to the driver is not relevant. For example, in Fig. 5(a) the building is highlighted as one of the features responsible for the image to be classified as a hotspot. In a way, it might be logical as multi-story buildings attract vehicles and pedestrians, leading to an accident. However, this building is not as useful as roads and roadsides to drivers for attentive driving. Therefore, to crop these features, we applied traditional image processing as graphically described in Fig. 5 along with the description of each sub-figure. Using traditional image processing techniques available in the OpenCV library, a pipeline to remove features selected by the CAM method in the upper part (300*640 size) of each original image (640*640 size) is made, because that part of the image hardly consists of roads and their related features. In the lower 280*640 part of each image, the features selected by the CAM method are surely related to the road and adjacencies and therefore should be selected by the system. Features contained in the region between the upper part (300*640) and lower part (280*640) of the image, i.e., 60*640 size, are selected, only if some part of the features also lie in the lower part. Therefore, in the image processing pipeline after thresholding the CAM map distance transform for generating a distance map in relevance to the object in question [47], and the watershed algorithm for segmentation purposes [48] are applied for feature selection. Finally, the contours are drawn according to the selected features on the original image. The contours are to be displayed in the HUD as a part of the visual notification.

### B. Attentive driving scheme based on visual notification

#### 1) Layout of visual information:

For visual notification, various types of HUDs have been deployed commercially. These systems span from augmented reality wearables to head-mounted displays and from full-fledged windshield HUD to a dashboard display [3, 4]. Wearable and head-mounted displays put excess pressure on the psychology of drivers as long-route drivers are not acclimated to it. On the other hand, the HDD forces drivers to remove their gaze from the road for looking at the small dashboard screen [3]. An HUD seems to be a viable solution for layout of visual notification information. However, the addition of a full windshield panel HUD can raise the price of a vehicle significantly. Also, for vehicles already in use, modification of a windshield HUD is a convoluted process.

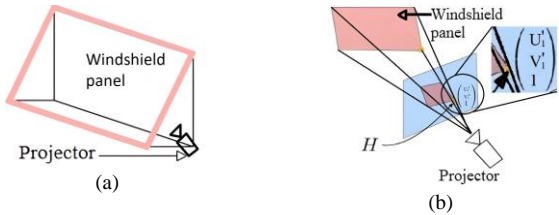

**Fig. 6.** (a) Projector/laser positioned for projecting on the windshield panel, and (b) usage of a homography matrix (H) for calculation of projection points due to center misalignment of the projector.

For approved mass adaptation, a simple low-cost real-time solution borrowing the design method from established projection systems like an electricity light machine organization (ELMO) projector is needed [49]. The ELMO projector uses the occlusion of light rays with translucent screens to fuse virtual and physical world images. Similar to the ELMO projector, a zoomable HUD with an integration of holographic and geometrical imaging has been proposed [50]. Based on the same methodology, the projector can be placed at one side of the front dashboard inside the vehicle, as shown in Fig. 6(a). Light from the projector gets occluded by a translucent windshield and the windshield acts as a projection screen.

However, the intensity of projecting light must be tuned such that the outside view is still clear to the driver. As the projector is not aligned to the center of the projecting screen, a homography matrix (see (H) in Fig. 6(b)) is calculated and used to find the correct projection area [51] according to the setup of different vehicle models.

#### 2) Visual notification system:

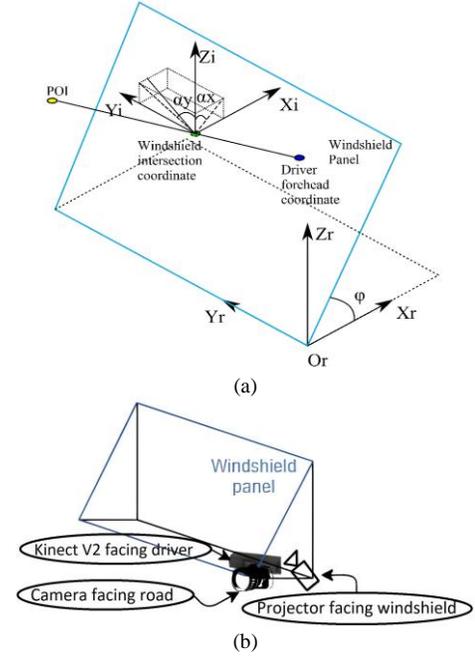

**Fig. 7.** (a) Coordinate geometry system presentation for POI (region containing AP features) and driver's forehead. (b) Hardware setup for the camera (bearing sensor), Kinect v2 (driver head coordinate sensing) and laser projection for the windshield as a notification screen.

Presentation of visual notifications based on screen placement has a vital impact on the psyche of the driver, and thus on the usability of the system. In this paper, a projection-based visual notification system is designed using the vehicle's windshield panel as a screen. The windshield glass obstructs the projected light and behaves as the notification screen. While driving, the gaze of the driver can be dynamic with focus adjustment to see different entities. Also, the driver has a degree of movement in the driving seat that allows for head movements. To analyze a complex road scene that is more susceptible to accidents, the driver's gaze changes rapidly while the head is comparatively stable; therefore, it is viable to only consider the driver's head movements and let the driver have the freedom to choose their gaze while checking for visual information on the projection screen. As the notification pattern, a windshield patch for POI is used, e.g., creating patches over AP features in hotspots. The windshield patch is defined as the contour drawn on the windshield panel at the point of intersection of the line joining the driver's forehead coordinate in the direction of the POI region. Given the geometry of the windshield, bearing angle of the POI, and forehead coordinate of the driver, as shown in Fig. 6 (a), the windshield patch intersection coordinate $(X_i, Y_i, Z_i)$ can be calculated with respect to reference coordinate $(X_r, Y_r, Z_r)$ and origin $(O_r)$, as presented in [4].

To find the patch on the windshield, the coordinates of the driver's forehead and bearing angles of the POI must be detected. To this end, a camera capable of depth-sensing, Kinect v2, is used. Kinect v2 is commercially available hardware that uses RGB cameras along with infrared projectors and detectors to measure depth by time of flight sensing technology for measuring back and forth time of receiving emitted IR light when scattered from an object. An implementation of an algorithm for forehead coordinate tracking is available at [52]. The bearing angle of the centroid of the areas with an AP feature must be calculated. For that, a camera capturing the road view is used as a bearing sensor with known camera parameters such as perspective angle and resolution at the plane coordinated by $x$ and $y$. Angles formed by the centroid of POI (POI in Fig. 7(a)) with respect to the camera, i.e., $\alpha x$ and $\alpha y$ in Fig. 7(a), are calculated as in [53, 54]. The hardware setup along with coordinate geometry is presented in Fig. 7. The camera and the Kinect v2 are assumed to be in the same position (see, Fig. 7(b)).

### 3) System Design:

In real-time, the system must access the vehicle GPS, the image from the camera, and the Kinect v2, while also having the accident hotspot database in the system's computer. The system checks the GPS data of the vehicle iteratively to identify the current location. The system checks if the vehicle is near a hotspot by comparing the distance between the current vehicle position and the location of hotspots stored in the database. If there is any hotspot within 200 meters of the vehicle's current location, the system will switch to the mode for AP features detection. Accordingly, the windshield patch for notification is calculated based on the centroid of the enclosed accident-feature as POI areas. Also, as training data captured the high variability of an urban view, classification of accident hotspots and non-hotspots for structurally similar urban cities will have similar efficacy as that of the New York city. This is verified in previous literature for different applications using street view images [15, 19].

## IV. SYSTEM EVALUATION AND ANALYSIS

### A. CNN classifier

#### 1) Environment setup:

To train the classifier, the entire dataset is randomly partitioned into training (70%), testing (20%), and validation (10%) subsets. Google street view images are resized to 224×224 pixels to match the input size of the pre-trained models used in the transfer learning process. In addition, normalization of the images is performed as per the requirements of pre-trained models. Each model is trained for 500 epochs with a batch size set to 8 in two possible training configurations for the analysis presented in TABLE I. In one configuration, the entire model is trained to re-calibrate the initial weights of the backbone CNN, while, for another, the backbone weights are fixed and only the weights of the last fully connected layer are trained. Binary cross-entropy is used as the loss function for training. The optimizer in the training process is the stochastic gradient descent (SGD) with momentum, using hyperparameters set as 0.001 for learning rate and 0.9 for momentum. The training was performed on a Windows machine configured with a GeForce RTX 1080 GPU and 32 GB RAM. The average training time for the models was around 4 hours.

#### 2) Training and model evaluation:

Four CNN models were used for the experiments: SqueezeNet, VGG16, ResNet-18, and DenseNet. In TABLE II, the accuracy, precision, recall, and F1 score are enlisted for two instances of training for each architecture. When all the weights of the CNN models are trained, specific AP features can be learned by the convolution layers. However, using only the last layer, which is the fully connected layer, the features used are similar to the ones learned for the Imagenet object classification dataset. Also, this accuracy difference in both training methods highlights the fact that AP features may not only be related to simple object classes but also the subtle complex linking of low-level features that occurs due to the interaction of different object classes in an image. Among all the CNN backbone models, the DenseNet (with all trainable weights) gives the best result. The models, VGG16 ResNet-18, DenseNet, and SqueezeNet, have a model size of 512 MB, 43MB, 27MB, and 2.82 MB, respectively. For vehicles with constrained memory capacity, SqueezeNet seems the most suitable, following Densenet. Also, the attention-based module with the CNN backbone, when trained with different configurations as listed in TABLE I, Densenet-ABM shows the highest accuracy. For identifying AP features using CAM methods, the trained CNN models in this section are utilized. The trained models can also be used for city-wide inspection to find accident-prone intersections and take proactive design measures accordingly [55].

TABLE II
EVALUATION METRICS FOR DIFFERENT CNN BACKBONE MODELS (FC = FULLY CONNECTED).

| CNN | | Accuracy | Precision | Recall | F1-Score |
|---|---|---|---|---|---|
| SqueezeNet | Only weights of the FC layer are trained. | 0.840 | 0.855 | 0.783 | 0.797 |
| VGG16 | | 0.819 | 0.821 | 0.808 | 0.791 |
| ResNet-18 | | 0.832 | 0.855 | 0.763 | 0.785 |
| DenseNet | | 0.819 | 0.847 | 0.771 | 0.783 |
| SqueezeNet | All the weights are trained. | 0.891 | 0.902 | 0.859 | 0.864 |
| VGG16 | | 0.904 | 0.916 | 0.888 | 0.887 |
| ResNet-18 | | 0.900 | 0.912 | 0.892 | 0.889 |
| DenseNet | | 0.905 | 0.903 | 0.876 | 0.877 |

Further analysis was performed by applying the trained model for a different urban city. Accident data collected for Austin Texas, US, from 2017 was used with 250,778 accident events. Clustering for finding hotspots and non-hotspots was performed similarly as for the New York City dataset. For the clusters obtained, 1470 images of hotspots and 1424 images of non-hotspots were collected using the Google Street View API. The results for the collected images were inferred using the CNN backbone classifier trained with New York City images. The highest accuracy obtained for the images of Austin is 0.758, as compared to 0.905 with the images of New York City, as shown in TABLE II. This difference in accuracy can be explained mainly by the variation in the cityscape of the cities of Austin and New York. The cityscape of Austin is mainly composed of open fields and green space, while the cityscape in New York is mainly composed of buildings and urban structures. Therefore, before applying a trained model for a target city, pre-analysis is required to verify that the training data is similar to the target cityscape. Nonetheless, the CNN backbone-based models trained with the New York city dataset demonstrated usefulness even with a city of a somewhat different cityscape.

### B. Accident-prone features visual sample analysis

A good explanation map of CAM for a given class should highlight the regions that are most relevant for the prediction of that class. For quantitative evaluation of the detected AP features, we studied the change in confidence between the setting when the full images are provided as input and when only images from which the explainable parts identified by CAM are cropped. It is expected that removing the explainable parts identified by CAM from the image will reduce the confidence of the model in its decision for that class when compared to its confidence when the full image is provided as input. The output of the Softmax layer in the model responsible for classifying a given class is used as the confidence metric. The reduction in confidence is expected because of both removing parts of the image (reducing the context) and removing explainable parts of an image for a given class. Therefore, the more the confidence "Change", the better. The confidence change metric is given as (10),

$$\text{Conf. Change Percent} = \left( \sum_{i=1}^{N} \frac{(Y_i^a - O_i^a)}{Y_i^a} \right) * 100 \qquad (10)$$

where $Y_i^a$ is the model's output score (confidence) for class 'a' on the $i^{\text{th}}$ image and $O_i^a$ is the same model's confidence in class 'a' giving an image as input from which the explainable parts are cropped. The image illustrations for different metrics are shown for a given sample in the supplementary material. This value is computed per image and averaged over the entire dataset. Therefore, for analysis, among the images of areas which are classified by the ABM models as the "hotspot" class, 70 images are randomly selected for further analysis. Details regarding the selected images are visualized in the supplementary material. For analysis, the accident-prone features identified by CAM methods as well as the image processing pipeline are cropped from the original images and the "Conf. Change percent" is obtained; therefore, the effect of accident-prone features on the result of classification can be evaluated. In the next stage of the analysis, the type of objects corresponding to AP features is obtained. From the result, the main type of objects acting as AP features are identified and analyzed.

The confidence $Y_i^a$ (in percent) is calculated and its average value is tabulated in the first column of TABLE III with the label "original". A similar metric as the one used in this work is also adopted in [56] for risk factors of injury severity of a driver. For understanding the relevance of the AP features selected by different CAM methods, $O_i^a$ (in percent) is calculated for images from which the selected highlighted parts of the CAM, after image processing, are cropped. The explainable parts are replaced by black color patches of the same shape. As expected, the value of $O_i^a$ is noted to be smaller than the value of $Y_i^a$. The results obtained with different combinations of three different CAM methods and models as CNN or ABM with different CNN backbones are tabulated in TABLE III.

### TABLE III
AVERAGE VALUE OF CONFIDENCE FOR DIFFERENT MODELS IN COMBINATION WITH DIFFERENT CAM METHODS AFTER APPLYING THE IMAGE PROCESSING PIPELINE (ROW-WISE LOWEST VALUES ARE HIGHLIGHTED IN BOLD FOR COLUMN 3,4, AND 5).

| Model | $Y_i^a$ - Original | $O_i^a$ - Grad CAM | $O_i^a$ - Grad CAM++ | $O_i^a$ - Score CAM | Conf. Change Percent |
|---|---|---|---|---|---|
| VGG | 98.15 | 82.33 | **79.8** | 86.72 | 18.7 |
| Resnet-18 | 94.63 | **82.04** | 82.44 | 84.2 | 13.3 |
| Squeezenet | *93.1* | 74.7 | **71.15** | 78.34 | *23.58* |
| Dense-net | 93.58 | 79.65 | **74.07** | 77.09 | 20.85 |
| VGG-ABM | 97.99 | 81.94 | **80.12** | 87.53 | 18.24 |
| Resnet-18-ABM | 95.58 | 83.73 | 82.87 | **81.27** | 14.27 |
| SqueezenetABM | *98.14* | **76.75** | 79.23 | 87.35 | *21.8* |
| Densenet-ABM | 69.32 | 54.72 | **54.17** | 56.82 | 21.86 |

### TABLE IV
AVERAGE VALUE OF CROPPED IMAGE PORTION WITH RESPECT TO ORIGINAL IMAGE AFTER APPLYING THE IMAGE PROCESSING PIPELINE (ROW-WISE LOWEST VALUES ARE HIGHLIGHTED IN BOLD FOR DIFFERENT CNN ARCHITECTURES).

| Model | GradCAM Feature Cropped | GradCAM++ Feature Cropped | ScoreCAM Feature Cropped |
|---|---|---|---|
| VGG | 11.23 | 12.06 | **7.36** |
| Resnet-18 | 16.72 | 19.77 | **15.46** |
| Squeezenet | 7.03 | *8.34* | **5.56** |
| Densenet | 17.98 | 24.20 | **17.45** |
| VGG-ABM | 10.90 | 11.78 | **4.86** |
| Resnet-ABM | 16.58 | 14.96 | **13.97** |
| Squeezenet-ABM | *9.61* | 10.25 | **8.43** |
| Densenet-ABM | **17.09** | 20.86 | 18.33 |

For the area selected by the CAM method, desired metrics include a high confidence score and a smaller highlighted area. The smaller highlighted area is desired as it attracts the driver's concentration on a specific area more than a wide area in the driver's field of view. The average fraction of the area covered by selected features of different combinations of CNN model and CAM method, after image processing, is presented in TABLE IV. The highest confidence (probability) change (23.58%) is for Squeezenet with GradCAM++ selecting 8.34% of the area to be cropped, on average (*Italic* in TABLE III and IV). However, if the confidence score is analyzed, it is seen that Squeezenet-ABM obtains 98.14% when compared to Squeezenet (93.1%). This shows that Squeeze-ABM produces more certain AP features. Also, Squeezenet-ABM along with GradCAM has a 9.61% cropped area on average with 21.8% confidence change (*Italic* in TABLE III and IV). Although for Squeezenet-ABM with GradCAM, when compared to Squeezenet with GradCAM++, the cropped area is greater with lesser confidence change, but a better confidence score leads to a more certain choice for AP feature selection. Based on this analysis, we suggest that the best combination of the model and CAM method is the Squeezenet-ABM with GradCAM.

Apart from the aforementioned metrics there are additional metrics prevalent in the literature when evaluating the output given by CAM [43]. To further evaluate the features detected by CAM, additional experiments were performed for the two best models (SqueezeNet with GradCAM++ and SqueezeNet-ABM with GradCAM). Results are shown in TABLE V. For Column 2 [37] in TABLE V, the metric is defined using the image with only explainable parts, as highlighted by CAM maps. If $E_i^a$ is the model's output score (confidence) for class 'a' on the i$^{th}$ image with the explainable part only, then the CAM confidence change (in percent) is given by (11). However, this metric is not highly relevant for AP feature evaluation, as these features are highly contextual (related to the full image) and cropping out the explainable part of the image with about 7-10% of the original image area will discard that contextual information. Therefore, to deal with contextual features, other metrics, as reported in Column 3 [43], 4 [59], and 5 [59] of TABLE V, have been used in previous studies. These metrics use different variations of images from which explainable parts are cropped. The cropped parts are replaced by different values rather than a black patch of the same shape. The details for the various changes for a given sample, as used in different metrics, can be seen in the supplementary material. Therefore, the confidence of these images with explainable parts cropped and filled with varied values, as per the details for these metrics, can be taken as $O'^a_i$ from the same model for class 'a'. After analyzing the values of different metrics in Columns 3, 4, and 5 of TABLE V, it can be seen that Squeezenet-ABM with GradCAM provides better values for the desired confidence change.

$$\text{CAM Confidence Change} = \left(\sum_{i=1}^{N} (E_i^c - Y_i^c)\right)100 \quad (11)$$

$$\text{Increase in Conf.} = 1 \text{ if } 0 < (E_i^c - Y_i^c), \text{ else } 0 \quad (12)$$

$$\text{Conf. Change} = \left(\sum_{i=1}^{N} (O'^c_i - Y_i^c)\right) \quad (13)$$

Further scene analysis for identifying different object types is done at the pixel level in the area highlighted by the CAM method. To detect the objects in the highlighted area, after cropping the area using an image processing pipeline, a DeepLabV3 model [57] pre-trained with the Cityscape dataset is used. There is a total of 19 possible object types for which pixel-wise labeling is performed. Considering portions of individual object types in the entire area of full images, the average percentage of the cropped area taken by roads, cars, sidewalks, buildings, sky, and vegetation are 38.1, 3.5, 2.0, 22.1, 24.2, and 5.3, respectively, and all other types contribute less than 1% each. In most cases, most of the pixels (regions) highlighted by the CAM methods are classified as roads, cars, and sidewalks. Since most traffic lights and the sky portion are located in the upper half of images, they are cropped by the image processing pipeline, thus leading to abrupt reduction of these objects.

After a manual inspection of the highlighted parts of images and the corresponding pixel-wise class (object) labeling in the original image, it is observed that road pixels are usually detected near crosswalks, near buildings, and in road areas prone to jaywalking, such as steep 'T' shaped intersections with a low field of view. Dynamic objects such as cars, buses, and trucks are observed especially in situations in which they occlude the critical field of view of the driver. From this context, a pedestrian or a bicycle appearing suddenly in front of the driver and dynamic objects in these intersections are usually AP features. Samples of the AP features identified with green contours by using Squeezenet-ABM with GradCAM are presented in Fig. 8. Also, all 70 images along with the contour and AP feature are visualized in the supplementary material for configuration (1) Squeezenet-ABM and GradCAM, and (2) Squeezenet and GradCAM++.

TABLE V
DIFFERENT METRIC FOR CAM MAP OF THE TWO BEST CONFIGURATIONS: (1) SQUEEZENET WITH GRADCAM++ (2) SQUEEZENET-ABM WITH GRADCAM. FOR THE METRIC IN THE SECOND COLUMN, LARGE VALUES ARE BETTER, AND FOR OTHER COLUMNS, A SMALLER VALUE IS BETTER. THE BEST VALUES ARE HIGHLIGHTED IN BOLD ACCORDINGLY.

| Configuration | Metric | CAM Confidence Change (%) (11), Increase in Conf. (12) | Conf. Change (13) for: Inv. CAM, Inv. CAM at 25%T | Conf. Change (13) for: ROAD at 25%T, ROAD at 10%T | Conf. Change (13) for: Avg {ROAD (20, 40, 60, 80)%T} |
|---|---|---|---|---|---|
| (1) | | **-10.2996, (1 is 15, 0 is 55)** | **-0.0956,** -0.2026 | -0.3176, -0.2700 | -0.3298 |
| (2) | | -13.9463 (1 is 7, 0 is 63) | -0.0923, **-0.2244** | **-0.4109** -0.2952 | **-0.3964** |

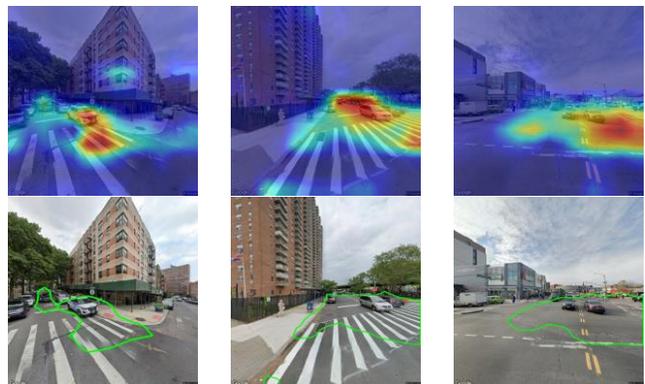

**Fig. 8.** Samples of the highlighted area using Squeezenet-ABM and GradCAM.

Next, to validate the proposed visual notification system, an analysis using saliency maps of each of the 70 images was made following the methodology presented in [39]. A saliency map shows where a person's gaze is mostly concentrated while looking the image of a road-view scene. The saliency mask is made by thresholding the saliency map at 100, in which the saliency map values vary from $0 - 255$. A high value means more saliency for that pixel. For CAM, an AP feature mask for the output after image processing pipeline is used. The visual saliency is calculated as the percent ratio of the common area of the saliency mask and the AP feature mask to area of saliency mask, as shown in (14)

$$\text{Visual Saliency (VS)} = \left(\sum_{i=1}^{N} \frac{(S_i^a \cap C_i^a)}{S_i^a}\right) * 100 \quad (14)$$

where $S_i^a$ and $C_i^a$ are the saliency mask area and the CAM AP feature mask for the i[th] image, respectively. The less visual saliency indicates that the AP features by the CAM have less saliency. In TABLE VI, visual saliency (VS) for different models and CAMs are presented. The least VS is for Squeezenet-ABM and GradCAM; therefore, this configuration is best for visual notification by highlighting the image areas which are accident-prone, but usually not focused on by human drivers.

TABLE VI
VISUAL SALIENCY FOR DIFFERENT MODELS AND CAMs. THE LOWEST VALUE IS HIGHLIGHTED IN BOLD.

| Model | VS for GradCAM Feature Cropped | VS for GradCAM++ Feature Cropped | VS for ScoreCAM Feature Cropped |
|---|---|---|---|
| VGG-ABM | 36.92 | 41.34 | 24.71 |
| Resnet-18-ABM | 46.59S | 48.61 | 44.98 |
| Squeezenet-ABM | **23.98** | 27.46 | 31.42 |
| Densenet-ABM | 40.23 | 53.38 | 47.31 |

### C. Simulation for analysis of layout for visual information

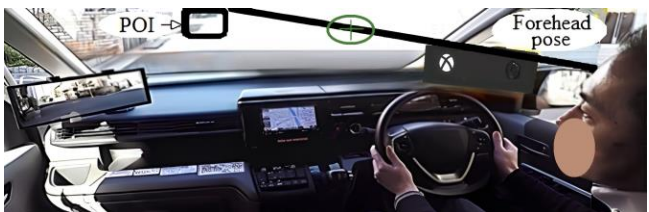

(a)

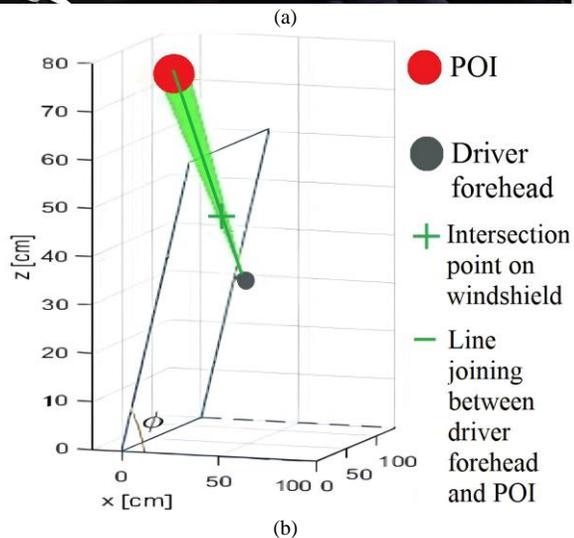

(b)

**Fig. 9.** (a) Layout of visual information. (b) MATLAB simulation for POI and driver forehead.

Given the driver forehead coordinates and the projection angle of the centroid of a POI (for a POI shown in Fig. 9(a, b)), windshield patch or point of intersection at windshield are studied using simulation by MATLAB. The POI, which corresponds to the forehead pose direction confined within the boundary of the windshield, is considered while the gaze of the driver outside the windshield is not. The setting in a real-world scenario is presented in Fig. 9(a) where the forehead pose direction (solid black line) is used to create green contours on the windshield with a green (+) sign as the intersection point of the line, between the forehead of the driver and the centroid of the POI, with the windshield. The forehead pose direction is calculated by using the line between the driver's forehead coordinates and the POI. Three scripts were made: first for cases when the driver's forehead position is fixed and the centroid of the POI position is mobile, second for the opposite of the first case, and third for when both are assumed to move. All variables are configurable in an input script, where the user can control the aspects of the simulation: the size of the windshield (L = width and h = height), the angle between the reference plan and the windshield plan (φ), the coordinates of the driver's forehead (D), and the centroid of the POI position (αx and αy). In the simulation, the coordinates of the centroid of the POI are assumed, as shown in Fig. 9(b) by a red circle. For calculating the intersection point at the windshield panel, only projection angles $(\alpha x, \alpha y)$ of the centroid of POI and the driver's forehead coordinates are used. For notifications using the projector, a homography matrix [51] is used to find the intersection point of the windshield with respect to the projector position.

## V. DISCUSSION AND CONCLUSION

This paper proposes an attentive driving system based on visual notification of accident-prone features. To identify accident-prone features, open data and driver-accessible dashcam video are used for real-time inference of a CNN model. CNN models combined with CAM methods are used to select accident-prone features in an image corresponding to an accident hotspot. Using a dataset created with Google street view images, CNN models can learn complex AP features for identifying an image as an accident hotspot with an accuracy up to 92%. These AP features, when identified by using CAM methods, give interpretability of decisions taken by the CNN model with a street view image representing the driver viewpoint. For improving AP features, an attention-based module is proposed that can provide better contextual features along with CNN. A sample analysis shows that removing 9.6%, on average, of the original image, which represents AP features obtained by Squeezenet-ABM (CNN model) with GradCAM (CAM method), causes up to 21.8% more probability of given area to be classified as a non-hotspot. For the visual information layout system, a concept that consists of an HUD located in the windshield panel of a vehicle is suggested.

As shown with the dataset of Austin Texas, USA, the CNN model trained with the dataset of the New York City can be applied to other urban cities with variability of classification performance allowed. Even in this case, given the variability of features in different cities, as future work it would be beneficial to have a global dataset that used to train a generalized classifier that can be used effectively for different cities. Additionally, for notification, AP features by the best configuration of Squeezenet-ABM and GradCAM have the least visual-saliency (VS), thus serving the purpose of visual notification the most, as it highlights usually not focused on by human drivers in road-view. Our framework relies on the AP features extracted from CAM methods. Given the recent methods proposed for post-hoc explainability it would be noteworthy to expand the analysis performed in this paper to include other explainable methods. Finally, the proposed framework relies on street view images collected under specific conditions. To improve generalization, the availability of data encompassing different conditions, such as weather and time of the day, would further improve the generalization of the proposed framework.

## ACKNOWLEDGMENT

This work was supported in part by the Institute of Information and Communications Technology Planning and Evaluation (IITP) Grant funded by Korea Government (MSIT) (Development of Artificial Intelligence Technology that Continuously Improves Itself as the Situation Changes in the Real World) under Grant 2020-000440.

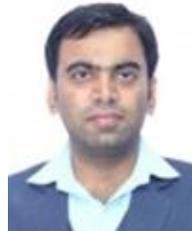

**Sumit Mishra** received B. Tech. in Electronics and Communication Engineering from Dr. A.P.J. Abdul Kalam Technical University, India. He is currently pursuing M.S. degree in Robotics from Korea Advanced Institute of Science and Technology, South Korea. Earlier, he worked with Learnogether Technologies Pvt. Ltd. as Research Consultant. His research interest is in deep learning, system modeling and design using multi-modal data.

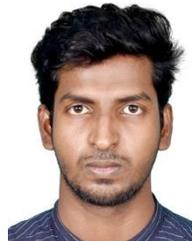

**Praveen Kumar Rajendran** received B.E. in Electrical and Electronics Engineering from Anna University, India. He is currently pursuing M.S. degree in Future Vehicle Program from Korea Advanced Institute of Science and Technology, South Korea. Earlier, he worked with SL Corporation as a Software Engineer. His research interests include deep learning, 3D computer vision and autonomous driving.

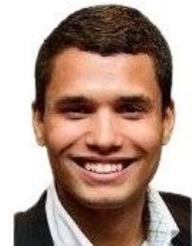

**Luiz Felipe Vecchietti** received the B.Sc. degree in electronics and computer engineering and the M.Sc. degree in electrical engineering from the Federal University of Rio de Janeiro, Rio de Janeiro, Brazil, in 2015 and 2017, respectively, and the Ph.D. degree in green transportation from the Korea Advanced Institute of Science and Technology, Daejeon, Korea, in 2021. His research interests include deep reinforcement learning, bioinformatics, digital signal processing, and applied deep learning.

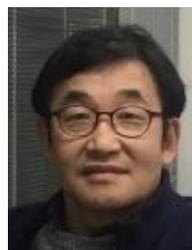

**Dongsoo Har** (Senior Member, IEEE) received the B.Sc. and M.Sc. degrees in electronics engineering from Seoul National University, and the Ph.D. degree in electrical engineering from Polytechnic University, Brooklyn, NY, USA. He is currently a Faculty Member of KAIST. He has authored and published more than 100 articles in international journals and conferences. He has also presented invited talks and keynote in international conferences. His main research interests include optimization of communication system operation and transportation system development with embedded artificial intelligence. He was a member of the Advisory Board, the Program Chair, the Vice Chair, and the General Chair of international conferences. He was a recipient of the Best Paper Award (Jack Neubauer Award) from the IEEE TRANSACTIONS ON VEHICULAR TECHNOLOGY in 2000. He is an Associate Editor of the IEEE SENSORS JOURNAL.